%% file: main.tex
\documentclass[10pt,twocolumn,letterpaper]{article}

\usepackage[pagenumbers]{iccv} 

\input{preamble}

\definecolor{iccvblue}{rgb}{0.21,0.49,0.74}
\usepackage[pagebackref,breaklinks,colorlinks,allcolors=iccvblue]{hyperref}
\usepackage[accsupp]{axessibility}  
\usepackage{fancyhdr}

\def\paperID{5} 
\def\confName{ICCV}
\def\confYear{2025}

\title{\selfadapt: Unsupervised Domain Adaptation of Cell Segmentation Models}

\author{
    Fabian H. Reith\textsuperscript{1,2,4,5}\thanks{Corresponding author: fabian.reith@mdc-berlin.de}\ ,
    Jannik Franzen\textsuperscript{1,3,4,5},
    Dinesh R. Palli\textsuperscript{1,6}, \\
    J. Lorenz Rumberger\textsuperscript{2,4,5}\thanks{Equal contribution; when citing, it is permitted to change author order; author order was determined at random.}\ ,
    Dagmar Kainmueller\textsuperscript{3,4,5}\footnotemark[2]
    \\[1.5em] 
    \textsuperscript{1}Charité - Universitätsmedizin, Berlin, Germany \quad
    \textsuperscript{2}Humboldt-Universität zu Berlin, Berlin, Germany \\
    \textsuperscript{3}Universität Potsdam, Digital Engineering Faculty, Potsdam, Germany \\
    \textsuperscript{4}Helmholtz Imaging \\
    \textsuperscript{5}Max Delbrück Center for Molecular Medicine in the Helmholtz Association, Berlin, Germany \\
    \textsuperscript{6}Ludwig-Maximilians-University, Munich, Germany
}

\begin{document}
\maketitle

\thispagestyle{fancy}
\fancyhf{} 
\renewcommand{\headrulewidth}{0pt} 
\lfoot{\footnotesize{© 2025 IEEE. Personal use of this material is permitted. Permission from IEEE must be obtained for all other uses, in any current or future media, including reprinting/republishing this material for advertising or promotional purposes, creating new collective works, for resale or redistribution to servers or lists, or reuse of any copyrighted component of this work in other works.}}

\input{sec/0_abstract}
\input{sec/1_intro}

\input{sec/2_method}
\input{sec/3_results}
\input{sec/4_conclusion}


{
    \small
    \bibliographystyle{ieeenat_fullname}
    \bibliography{main}
}


\end{document}

%% file: preamble.tex
\usepackage[T1]{fontenc}
\usepackage{verbatim}
\usepackage{silence}
\let\vec\relax
\usepackage{booktabs}
\usepackage{makecell}
\usepackage{xcolor}     
\usepackage{pifont}     
\usepackage[misc]{ifsym}

\newcommand{\cmark}{\textcolor{green!50!black}{\checkmark}}
\newcommand{\xmark}{\textcolor{red!60!black}{\textbf{\ding{55}}}}

\newcommand{\selfadapt}{SelfAdapt}



%% file: sec/0_abstract.tex

\begin{abstract}
Deep neural networks have become the go-to method for biomedical instance segmentation. Generalist models like Cellpose demonstrate state-of-the-art performance across diverse cellular data, though their effectiveness often degrades on domains that differ from their training data. While supervised fine-tuning can address this limitation, it requires annotated data that may not be readily available. We propose \selfadapt, a method that enables the adaptation of pre-trained cell segmentation models without the need for labels. Our approach builds upon student-teacher augmentation consistency training, introducing L2-SP regularization and label-free stopping criteria. We evaluate our method on the LiveCell and TissueNet datasets, demonstrating relative improvements in AP$_{0.5}$ of up to 29.64\% over baseline Cellpose. Additionally, we show that our unsupervised adaptation can further improve models that were previously fine-tuned with supervision. We release \selfadapt\ as an easy-to-use extension of the Cellpose framework. The code for our method is publicly available at \url{https://github.com/Kainmueller-Lab/self_adapt}.
\end{abstract}

%% file: sec/1_intro.tex
\section{Introduction}
\label{sec:intro}

Deep neural networks have become a widely applied approach for cell instance segmentation in microscopy data. In particular, generalist models such as Cellpose \cite{Stringer2021-wu}, StarDist \cite{Schmidt2018-cq}, Mesmer \cite{greenwald2022whole} and Micro-SAM \cite{Archit2023-wn} have emerged as powerful segmentation tools, capable of handling a broad range of imaging modalities. These models are trained on diverse datasets and achieve state-of-the-art generalization performance. Nevertheless, their performance still often drops significantly on domains which differ from their training data \cite{Shi2023-lh,Pachitariu2022-eo}. This limitation is particularly pronounced in biomedical imaging, where new data is often "out-of-domain" due to the wealth of cell types and tissues considered in biomedical research, highly diverse protocols for sample preparation, and continuously evolving imaging modalities -- with potentially significant impact on model performance.

To address performance degradation caused by domain shifts, domain adaptation (DA) methodology seeks to adapt a model trained on a source domain to an unlabeled or sparsely labeled target domain. 
Traditional approaches for domain adaptation rely on supervised fine-tuning on the target domain \cite{Pachitariu2022-eo}, autoencoder pre- or auxiliary training \cite{ghifary2016deep,hoffman2016fcns,roels2019domain} on source and target domain, and domain adversarial learning \cite{ganin2016domain,hoffman2016fcns}. However, these approaches require either labeled data from the target domain, or labeled data from the source domain, or at least image data from the source domain. This makes them unsuitable for adapting  off-the-shelf models: First, acquiring labeled target annotations in biomedical applications is often prohibitively expensive and time-consuming, requiring expert knowledge and extensive resources; Second,  off-the-shelf models usually do not ship with their training data. 

This challenge is addressed by source-free unsupervised domain adaptation (UDA), which enables adaptation without any access to source data nor target labels. 
In particular, student-teacher frameworks with exponential moving average (EMA) updates \cite{Tarvainen2017-wu} have shown strong potential for improving model generalization. These methods maintain a teacher model that is iteratively updated via EMA and used to generate pseudo-labels for a student model, allowing the network to learn from unlabeled samples. Techniques like FixMatch \cite{Sohn2020-ij} and ACTIS \cite{Rumberger2023-am} further enhance this process by enforcing consistency between a student’s predictions under different augmentations, extracting reliable training signals from unlabeled data. While originally showcased in semi-supervised scenarios where the UDA objective is complemented by small sets of labeled target data, Archit and Pape \cite{Archit2023-wn} show that these approaches can be adapted for pure source-free UDA scenarios in bio-medical applications; However, to do so, their method relies on a specialized architecture for uncertainty quantification \cite{kohl2018probabilistic} and thus does not lend itself to off-the-shelf pre-trained models. 

To fill this gap, we propose \selfadapt, a source-free UDA framework whose core training procedure is applicable to any model trained for pixel-wise classification- and/or regression tasks. We showcase this framework on cell instance segmentation, for which we also introduce and evaluate two distinct label-free stopping criteria: a general, representation-based metric (embedding distance) and an instance-specific metric (False Negative rate). 
\selfadapt\ expands over ACTIS \cite{Rumberger2023-am}, which only considered models trained for pixel-wise classification tasks. By incorporating regression targets, \selfadapt\ becomes applicable to many popular cell segmentation models, including Cellpose \cite{Stringer2021-wu,Pachitariu2022-eo}, Stardist \cite{Schmidt2018-cq}, and Mesmer \cite{greenwald2022whole}. We further expand over ACTIS by introducing L2-SP regularization \cite{Li2018-mk} into the objective: L2-SP regularization constrains the adapted weights to stay close to their pre-trained values, thus serving to replace the regularizing effect of ACTIS's supervised component without compromising training stability.

We showcase \selfadapt\ by applying it to Cellpose, arguably the most widely used off-the-shelf cell instance segmentation model, improving over the generalist baseline by large margins. Interestingly, our results even suggest improvements over off-the-shelf \emph{fine-tuned} Cellpose models, i.e., models adapted to the target domains in a traditional supervised fashion \cite{Pachitariu2022-eo}.

In summary, our key contributions are:
\begin{enumerate}
    \item \selfadapt, an easy-to-use, label-free UDA method featuring a broadly applicable L2-SP-regularized training procedure and two distinct early stopping criteria: a general embedding-based metric and another tailored for instance-based tasks.
    \item Seamless integration of \selfadapt\ into the Cellpose framework for immediate deployment to the community.
    \item Comprehensive evaluation on multiple cellular imaging datasets, demonstrating substantial performance improvements over generalist as well as fine-tuned Cellpose baselines.
\end{enumerate}

\begin{figure*}[t]
\centering
\includegraphics[width=\textwidth]{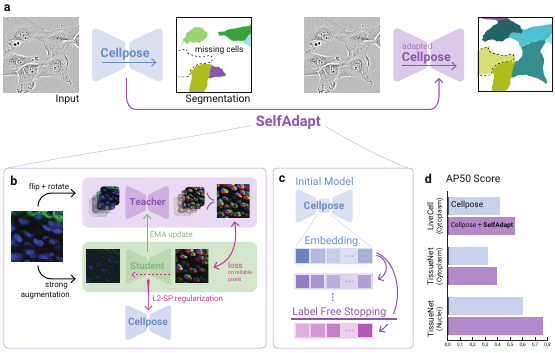}
\caption{Overview of \selfadapt. a) Generalist Cellpose model showing limited performance on a target domain, while our adapted model improves segmentation accuracy. b) Student-teacher framework with strong augmentations for the student and weak augmentations (flip+rotate) for the teacher, coupled with EMA updates and L2-SP regularization. c) Early stopping mechanism based on embedding distances from initial model. d) Performance comparison showing AP$_{0.5}$ improvements across datasets.}
\label{fig:method}
\end{figure*}

%% file: sec/2_method.tex
\section{Method}
\label{sec:method}

\subsection{Augmentation Consistency Training}
\selfadapt\ builds upon ACTIS \cite{Rumberger2023-am}, a student-teacher framework \cite{Tarvainen2017-wu} that enforces augmentation consistency, i.e., consistency between predictions on strongly and weakly augmented views of the input data, and employs teacher-generated pseudo labels \cite{Sohn2020-ij,Lee2013-uf}. ACTIS further advances this training framework for cell instance segmentation by introducing a confidence-based loss masking scheme that filters uncertain predictions, combined with temporal ensembling \cite{Laine2016-gz} that improves pseudo-label quality through prediction averaging.

Following ACTIS, \selfadapt's teacher model generates pseudo-labels by averaging predictions across multiple weakly augmented versions of an input image. Specifically, we apply flips and 90-degree rotations as weak augmentations, compute predictions for each transformation, and apply inverse transformations before averaging the outputs. 

We further follow ACTIS in their confidence-based loss masking scheme, which applies to pixel-wise classification outputs. Expanding upon ACTIS, we also tackle pixel-wise regression outputs, using standard deviation across the different teacher predictions, i.e., a standard measure for regression uncertainty, as filtering criterion. For each scalar regression target, we maintain a moving average of the 80th uncertainty percentile to determine reliable regions. The student model, receiving strongly augmented inputs with intensity adjustments, contrast changes, blurring and Gaussian noise, is trained using a weighted combination of mean squared error for regression targets and cross-entropy loss for classification targets:
%
\begin{equation}
\begin{split}
\mathcal{L}_{\text{self}} ={}& \mathcal{L}_{\text{CE}}(p_{\text{student}}, p_{\text{teacher}}) \cdot M_p \\
                            & + \lambda_{bal} \cdot \mathcal{L}_{\text{MSE}}(f_{\text{student}}, f_{\text{teacher}}) \cdot M_f\,,
\end{split}
\end{equation}
where $p$ represents class probability maps, f represents regression output predictions, $M_p$ and $M_f$ are binary masks excluding pixels with uncertainty above the threshold, and $\lambda_{bal}$ balances the contribution of regression- vs.\ classification outputs. Only pixels with uncertainty below our dynamic threshold contribute to the loss, thus steering training to focus on reliable regions.

\subsection{L2-SP Regularization}
To preserve the features learned during pre-training while allowing for domain-specific adaptations, we employ L2-SP  regularization~\cite{Li2018-mk} alongside standard weight decay. The L2-SP regularization term is defined as:
\begin{equation}
    \mathcal{L}_{\text{L2-SP}} = \lambda_{L2\text{-}SP} \sum_i (w_i - w_i^0)^2\,,
\end{equation}
where $w_i$ represents the current weights, $w_i^0$ denotes the initial pre-trained weights, and $\lambda_{L2\text{-}SP}$ controls the regularization strength. This term explicitly encourages the model to maintain proximity to its initialization, serving to stabilize training in place of the supervised sub-objective of ACTIS. 

\subsection{Label-free Early Stopping Criteria}\label{sec:stopping}

In fully unsupervised learning scenarios, determining the optimal stopping point is a critical challenge. While manual inspection is an option, it is subjective and not scalable. To address this, we propose an approach that relies on two label-free metrics to signal when to stop training.

The first is the False Negative (FN) rate, adapted from~\cite{Ori-Press2024-ym}, which tracks the fraction of instances detected by the initial model ($\theta_0$) that are missed by the current model ($\theta_t$). It is defined as:
\begin{equation}\label{eq:fn_rate}
\text{FN}_{\text{rate}}(t) = \frac{\text{FN}(t)}{\text{TP}(t) + \text{FN}(t)}.
\end{equation}
Here, $\text{TP}(t)$ are instances detected by both the initial and current models, while $\text{FN}(t)$ are instances detected initially but missed by the model at iteration $t$.

The second is the mean Euclidean distance ($D_{emb}$) between the current and initial bottleneck feature embeddings ($E$) across the validation set:
\begin{equation}\label{eq:emb_dist}
D_{emb}(t) = \frac{1}{N} \sum_{i=1}^{N} ||E(x_i; \theta_t) - E(x_i; \theta_0)||_2.
\end{equation}
In this formulation, $E(x_i; \theta_t)$ is the bottleneck feature embedding for an input image $x_i$ from the model with its current weights, which is compared against $E(x_i; \theta_0)$, the corresponding embedding from the initial, pre-trained model. This distance is then averaged over all $N$ images in the validation set.

We selected these two metrics after a systematic investigation of a broader set of candidates, including other output-based metrics (e.g., prediction confidence, uncertainty) and representation-based metrics (e.g., proximity to source embeddings). A detailed analysis comparing these candidates and justifying our selection is presented in the Results section (Sec.~\ref{sec:stopping_analysis}).

Finally, the two selected criteria were calibrated on a representative dataset to establish general-purpose thresholds, which are used in all subsequent experiments.

%% file: sec/3_results.tex
\section{Results}
\label{sec:results}

\begin{figure*}[t]
\centering

\includegraphics[width=0.75\textwidth]{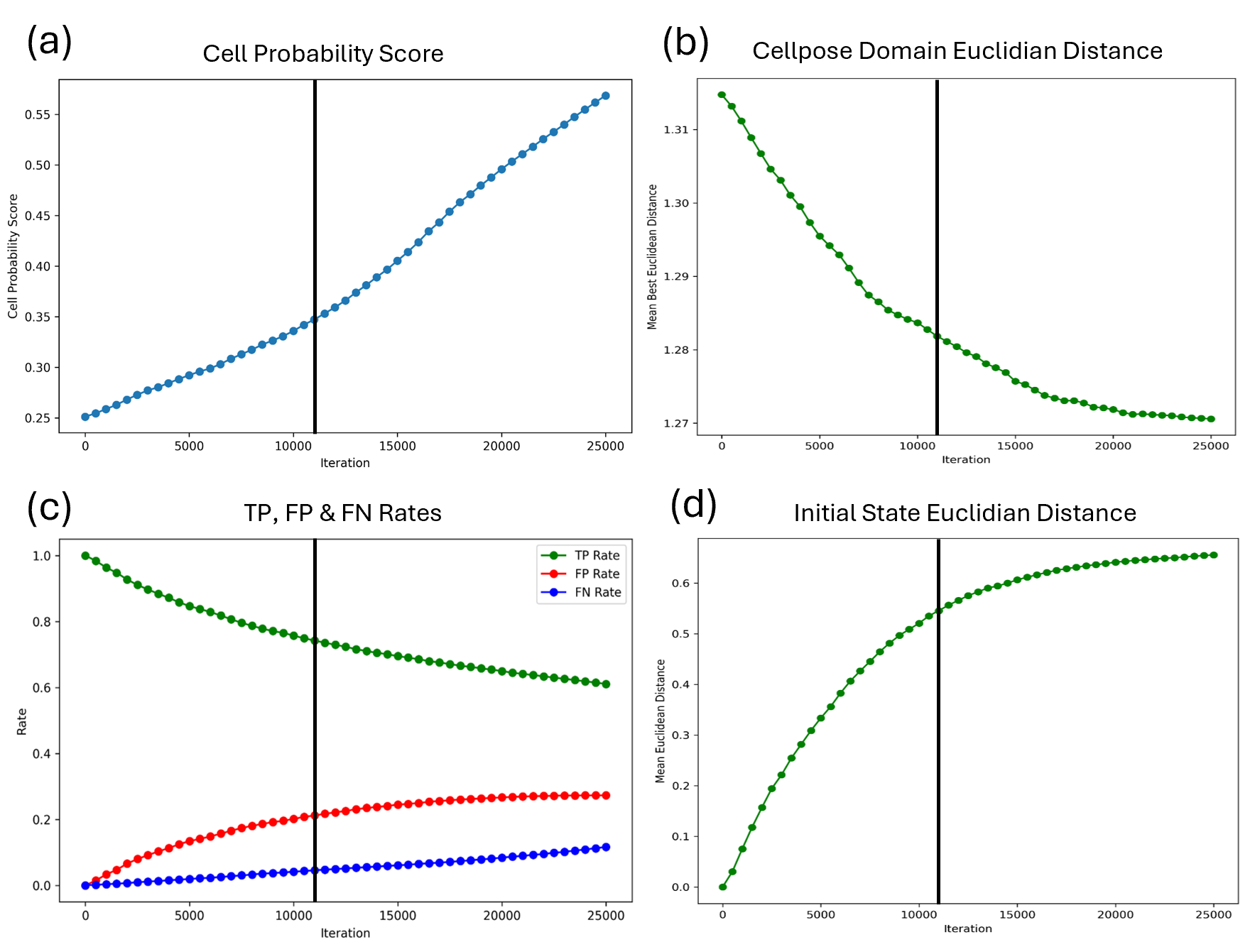}
\caption{Analysis and selection of label-free early stopping criteria. We compare candidate metrics on a representative adaptation run (TissueNet Nuclei). The vertical black line indicates the iteration with the peak "oracle" performance on the test set. 
(a, b) Many intuitive metrics are unsuitable. Prediction confidence (a) increases monotonically as the model overfits, while proximity to the source domain (b) offers no clear stopping point and violates our source-free requirement. 
(c, d) In contrast, our chosen metrics provide a reliable signal. The drift in instance detection, measured by the FN Rate (c, blue line), and the drift in feature space, measured by the embedding distance $D_{emb}$ (d), both produce monotonic curves with a clear, steady trend, making them ideal for thresholding.}
\label{fig:stopping_analysis}
\end{figure*}

\begin{table*}[t] 
\caption{Performance comparison on LiveCell and TissueNet datasets. Our self-adaptation method improves both base and fine-tuned models, using three stopping criteria: False Negatives rate (FN), Embedding distance (Emb), and test-based maximum. Results show AP$_{0.5}$ scores with standard deviations over three runs. For TissueNet (Nuclei), no fine-tuned model is available by~\cite{Pachitariu2022-eo}}
\centering
\label{tab:main_results}
\begin{tabular}{lccc}
\toprule
Model & \makecell{LiveCell\\(Cytoplasm)} & \makecell{TissueNet\\(Cytoplasm)} & \makecell{TissueNet\\(Nuclei)} \\
\midrule
Cellpose base model & .415 & .324 & .603 \\
+ \selfadapt (Early Stop: FN) & .480 $\pm$ .008 & .394 $\pm$ .014 & .759 $\pm$ .001 \\
+ \selfadapt (Early Stop: Emb) & .503 $\pm$ .003 & .391 $\pm$ .013 & .754 $\pm$ .003 \\
+ \selfadapt (Test Max) & .538 $\pm$ .002 & .395 $\pm$ .015 & .760 $\pm$ .001 \\
\midrule
Cellpose fine-tuned model & .695 & .750 & n/a \\
+ \selfadapt (Early Stop: FN) & .693 $\pm$ .004 & .752 $\pm$ .001 & n/a \\
+ \selfadapt (Early Stop: Emb) & .704 $\pm$ .001 & .755 $\pm$ .000 & n/a \\
+ \selfadapt (Test Max) & .705 $\pm$ .002 & .763 $\pm$ .000 & n/a \\
\bottomrule
\end{tabular}
\end{table*}

\begin{figure*}[t]
\centering
\includegraphics[width=\textwidth]{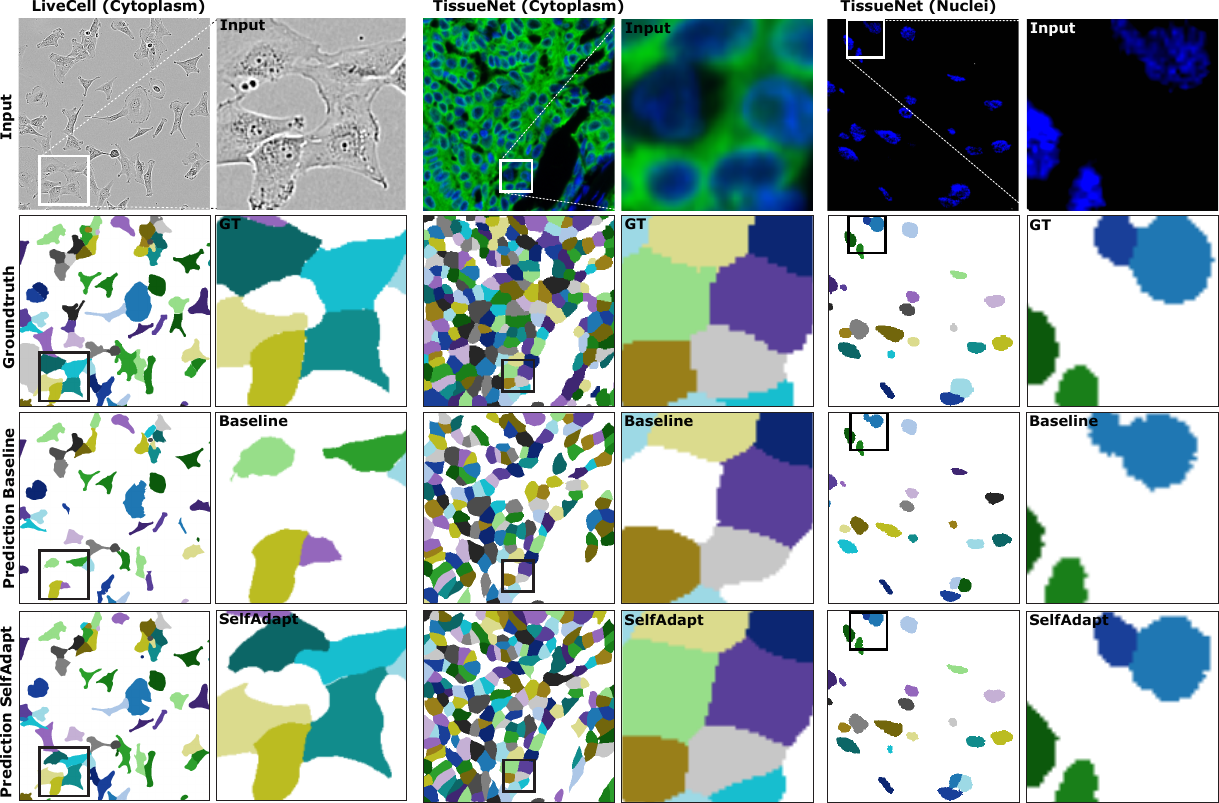}
\caption{Qualitative comparison of segmentation results. Each column shows results on a different dataset: LiveCell (Cytoplasm), TissueNet (Cytoplasm), and TissueNet (Nuclei). For each dataset, we show the input image (top), ground truth segmentation (second row), baseline Cellpose predictions (third row), and predictions after self-adaptation (bottom row). Insets highlight cases where self-adaptation improves detection of previously missed cells (LiveCell, TissueNet Cytoplasm) and better separation of adjacent cells (TissueNet Nuclei).}
\label{fig:qualitative}
\end{figure*}

\begin{table*}[t]
\caption{Ablation study examining the impact of key components in our method. Removing student augmentations or L2-SP regularization significantly impacts performance, while other components show more moderate effects.}
\centering
\label{tab:ablation}
\begin{tabular}{lccccccc}
\toprule
& \multicolumn{4}{c}{Ablations} & \multicolumn{3}{c}{AP$_{0.5}$ split by dataset}\\
\cmidrule(lr){2-5}\cmidrule(lr){6-8}
& \makecell{Confidence \\ Filtering}
& \makecell{Teacher \\ TTA}
& \makecell{L2-SP \\ Loss}
& \makecell{Student \\ Augmentations}
& \makecell{LiveCell\\(Cytoplasm)}
& \makecell{TissueNet \\ (Cytoplasm)} 
& \makecell{TissueNet \\ (Nuclei)} \\
\midrule
& \cmark & \cmark & \cmark & \cmark 
& .538 ± .002 & .395 ± .012 & .760 ± .001 \\
& \xmark & \cmark & \cmark & \cmark
& .537 ± .004 & .393 ± .006 & .760 ± .001 \\
& \cmark & \xmark & \cmark & \cmark
& .536 ± .001 & .395 ± .013 & .759 ± .001 \\
& \cmark & \cmark & \xmark & \cmark 
& .528 ± .005 & .391 ± .021 & .642 ± .009 \\
& \cmark & \cmark & \cmark & \xmark
& .451 ± .005 & .357 ± .004 & .615 ± .002 \\
\bottomrule
\end{tabular}
\end{table*}

\subsection{Experimental Setup}
\label{sec:exp_setup}

We evaluate \selfadapt\ by adapting state-of-the-art Cellpose models on three challenging benchmarks: cytoplasm segmentation on the LiveCell dataset~\cite{Edlund2021-mh}, and both cytoplasm and nuclei segmentation on TissueNet~\cite{greenwald2022whole}. To ensure fair comparison, we train on the official training set and report performance on the test set for each benchmark. All experiments are run three times with different random seeds, and we report the mean and standard deviation of the Average Precision at an IoU of 0.5 (AP$_{0.5}$) \cite{maier2024metrics}. We selected this metric as it is a common standard for cell instance segmentation and the primary metric used in the original Cellpose papers \cite{Stringer2021-wu,Pachitariu2022-eo}, ensuring a direct and fair comparison with the baseline models.

Our implementation details are as follows: We train for 25,000 iterations with a batch size of eight. The AdamW~\cite{Loshchilov2017-ww} optimizer is used with a weight decay of 0.001. The learning rate follows a linear warm-up and a cosine annealing schedule. For our loss function, we set the balancing weight $\lambda_{bal}=0.5$. The L2-SP regularization strength, $\lambda_{L2\text{-}SP}$, was tuned based on the specific model-dataset combination. For the results in Table~\ref{tab:main_results}, the values were set as follows: a strong regularization of $\lambda_{L2\text{-}SP} = 10^{-2}$ was used for the TissueNet (Nuclei) model and the fine-tuned TissueNet (Cytoplasm) model. A moderate value of $10^{-4}$ was used for the base models on LiveCell and TissueNet (Cytoplasm). Finally, a weaker regularization of $10^{-5}$ was applied when adapting the fine-tuned LiveCell model. The teacher model is updated via EMA with $\alpha=0.99$. For cytoplasm and nuclei segmentation, we use the generalist 'cyto2' and 'nuclei' Cellpose models, respectively. For experiments on already fine-tuned models, we use the domain-specific checkpoints provided by Cellpose 2.0~\cite{Pachitariu2022-eo}.

\subsection{Analysis and Selection of Early Stopping Criteria}
\label{sec:stopping_analysis}

As outlined in Sec.~\ref{sec:stopping}, our method relies on automated, label-free criteria to determine when to stop adaptation. To justify our choice, we performed a detailed analysis on a representative adaptation run (TissueNet Nuclei), comparing numerous candidate metrics against the true test set performance (the "oracle"). Our findings are summarized in Figure~\ref{fig:stopping_analysis}.

We first investigated metrics derived directly from the model's output. A naive approach is to monitor prediction confidence, such as the mean cell probability. However, this proved highly unreliable (Fig.~\ref{fig:stopping_analysis}a). We observed that confidence increased monotonically throughout training, continuing to rise even as the true performance degraded. We also explored more sophisticated uncertainty measures, such as the variance in predictions across multiple test-time augmentations (TTA). While these metrics were not always monotonic, their peaks and troughs did not correlate with the oracle performance peak, often providing a misleading signal for when to stop.

Next, we evaluated metrics based on the model's internal feature representations. One hypothesis was that the adapted target embeddings should move closer to the general feature space of the source domain. As shown in Fig.~\ref{fig:stopping_analysis}b, this proximity (measured by Euclidean distance) does indeed provide a smooth monotonic signal. However, this approach has two critical drawbacks: it violates our core source-free constraint by requiring access to the original training data, and as we will show, a source-free alternative provides an equally reliable signal.

This analysis led us to select two metrics that are both source-free and reliably track model drift from its stable, pre-trained state. The first is the False Negative (FN) rate (Fig.~\ref{fig:stopping_analysis}c), a self-consistency metric that measures the fraction of initially detected instances that the model "forgets" over time. The second is the mean Euclidean distance, $D_{emb}$ (Fig.~\ref{fig:stopping_analysis}d), which measures the drift of the model's feature embeddings from their robust initial state. Both of these metrics provide predictable, monotonic curves that are ideal for setting a threshold, without needing access to labels or source data. We note that while the FN rate is effective, it is most reliable when the initial model does not suffer from severe over-segmentation; in such cases, the instance-independent $D_{emb}$ metric provides a more robust signal.

Based on this comprehensive analysis, we calibrated general-purpose thresholds of $\tau_{FN} = 0.05$ and $\tau_{emb} = 0.5$. These fixed, pre-calibrated thresholds are used in all subsequent experiments to demonstrate a truly automated adaptation pipeline.

\subsection{Adaptation Performance}

Table~\ref{tab:main_results} presents the main quantitative results of our method. On all three benchmarks, the baseline Cellpose models show a significant performance drop compared to supervised models. Our method, \selfadapt, substantially closes this gap. When using the oracle "Test Max" as an upper bound, \selfadapt\ improves the AP$_{0.5}$ by relative margins of 29.6\% on LiveCell, 21.9\% on TissueNet (Cytoplasm), and 26.0\% on TissueNet (Nuclei).

Crucially, our automated label-free stopping criteria, using fixed thresholds calibrated on a separate representative run, capture a large portion of these potential gains. We validated this by applying the fixed thresholds ($\tau_{FN}=0.05$, $\tau_{emb}=0.5$) across all test runs. On the TissueNet (Nuclei) dataset, the same domain used for calibration, the FN and Embedding Distance criteria achieve 99.4\% and 96.2\% of the maximum possible improvement, respectively. 

More importantly, these fixed thresholds generalize remarkably well to entirely new datasets. For TissueNet (Cytoplasm), the FN and Embedding Distance criteria capture 98.6\% and 94.4\% of the maximum possible gain. For the visually distinct LiveCell dataset, the Embedding Distance criterion is particularly effective, achieving 71.5\% of the maximum possible improvement, while the FN criterion still captures a substantial 52.8\%. This demonstrates that our pre-calibrated thresholds can be used "off-the-shelf" to effectively adapt models to new domains without any labels or further tuning.

Interestingly, \selfadapt\ can even further refine models that have already been fine-tuned with supervision (Table~\ref{tab:main_results}, bottom). For example, the Embedding Distance criterion further improves the fine-tuned LiveCell model, capturing 90.0\% of the small remaining performance gap. While this demonstrates the robustness of our method, we note that in a supervised fine-tuning scenario, labeled validation data would typically be available, making automated label-free stopping less critical.

Figure~\ref{fig:qualitative} provides a qualitative view of these improvements. \selfadapt\ enables the detection of cells missed by the baseline model and improves the separation of adjacent cells that were previously merged, corroborating the quantitative gains.

\subsection{Ablation Study}

To understand the contribution of each component of \selfadapt, we conducted an ablation study (Table~\ref{tab:ablation}). The results highlight two critical components: strong student augmentations and L2-SP regularization. Removing student augmentations leads to a collapse in performance, as the student-teacher framework has no meaningful discrepancy to learn from. Removing L2-SP regularization is also highly detrimental, especially for the TissueNet (Nuclei) dataset, where it causes a 15.5\% relative performance decrease. This confirms that constraining the model to stay close to its powerful initial weights is key for stable unsupervised adaptation. In contrast, confidence filtering and teacher test-time augmentation (TTA), while beneficial, have a more moderate impact.

%% file: sec/4_conclusion.tex
\section{Conclusion}
\label{sec:conclusion}

We presented a practical method for unsupervised domain adaptation of cell segmentation models that enables immediate adaptation to new domains without requiring labels. Our approach combines augmentation consistency-based training with L2-SP regularization, demonstrating substantial improvements over baseline performance across multiple cellular image segmentation tasks. The success of L2-SP regularization suggests that for powerful generalist models like Cellpose, explicitly preventing the model from drifting too far from its robust initial weights is a key factor for stable adaptation, especially when learning from noisy pseudo-labels. The method proves effective not only for adapting base models but also for further improving supervised fine-tuned models.

\selfadapt\ features a default, "off-the-shelf" early stopping criterion as determined via our experiments. At the same time, users can also leverage manual early stopping / snapshot selection and re-use respective inferred stopping criteria as they see fit. 

We ship \selfadapt\ as an easy-to-use extension of the Cellpose framework, requiring minimal setup while maintaining full compatibility with Cellpose's extensive functionality. This contribution aims to make domain adaptation more accessible to the biomedical research community, enabling improved cell segmentation across diverse imaging conditions without the need for additional annotations.

\section*{Acknowledgments}
Funding: German Research Foundation (DFG) Research Training Group CompCancer (RTG2424); Synergy Unit of the Helmholtz Foundation Model Initiative.
%